\documentclass[preprint,5p,times,twocolumn]{elsarticle}

\usepackage{times}
\usepackage{epsfig}
\usepackage{graphicx}
\usepackage{amsmath}
\usepackage{adjustbox}
\usepackage{pifont}
\usepackage{caption}
\usepackage{subcaption}
\usepackage{xspace}
\usepackage{comment}
\usepackage{amssymb}

\usepackage{amsmath}
\usepackage{graphicx}
\usepackage{tabularx}
\usepackage{longtable}
\usepackage{lscape}
\usepackage{multirow}
\usepackage{bbding}
\usepackage{makecell}
\usepackage{url}
\usepackage{color, colortbl}

\usepackage{booktabs, multirow}
\usepackage{adjustbox}

\usepackage{outlines}
\usepackage{amsmath}

\usepackage[colorinlistoftodos]{todonotes}

\reversemarginpar
\usepackage{pdfcomment}

\newcommand{\quotes}[1]{``#1''} 

\usepackage{soul}

\usepackage{adjustbox}

\soulregister{\quotes}{1}
\soulregister{\ref}{7}
\soulregister{\cite}{7}
\soulregister{\citep}{7}
\soulregister{\citet}{7}
\soulregister{\subsection}{1}
\soulregister{\section}{1}
\soulregister{\label}{1}
\soulregister{\futurelet}{0}
\soulregister{\textbf}{1}

\ifdefined\citep
\else
  \newcommand{\citep}[1]{(\cite{#1})}
\fi

\usepackage{amsfonts}
\definecolor{mypurple}{rgb}{0.85,0.7,1}
\definecolor{mygray}{rgb}{0.85,0.85,0.85}
\definecolor{mygreen}{rgb}{1,0.65,0.65}
\definecolor{mysand}{rgb}{0.9,0.8,0.5}

\usepackage{colortbl}
\definecolor{Gray}{gray}{0.9}

\usepackage{xspace}
\newcommand*{\eg}{e.g.\@\xspace}
\newcommand*{\epickitchens}{EPIC-Kitchens-100\@\xspace}
\newcommand*{\ek}{EK100\@\xspace}
\newcommand*{\ie}{i.e.\@\xspace}

\definecolor{up_color}{RGB}{204,51,0}
\definecolor{down_color}{RGB}{202,195,121}

\definecolor{cadmiumgreen}{rgb}{0.0, 0.42, 0.24}
\newcommand{\up}{{\color{cadmiumgreen}$\blacktriangle$}\hspace{.3em}}

\newcommand{\cmark}{\ding{51}}%

\newcommand{\ours}{EgoZAR\xspace}
\newcommand{\pushright}[1]{\ifmeasuring@#1\else\omit\hfill$\displaystyle#1$\fi\ignorespaces}
\newcommand{\pushleft}[1]{\ifmeasuring@#1\else\omit$\displaystyle#1$\hfill\fi\ignorespaces}

\makeatletter
\def\ps@pprintTitle{%
  \let\@oddhead\@empty
  \let\@evenhead\@empty
  \def\@oddfoot{\reset@font\hfil\thepage\hfil}
  \let\@evenfoot\@oddfoot
}
\makeatother

\usepackage{url}
\usepackage{xcolor}
\usepackage{tikz}
\usepackage{pgfplots}
\pgfplotsset{compat=1.18}

\journal{ArXiV}

\begin{document}

\begin{frontmatter}

\title{\textbf{Egocentric zone-aware action recognition across environments}}

\author{Simone Alberto Peirone\corref{cor}\fnref{fn1}}\cortext[cor]{These authors contributed equally.}\ead{simone.peirone@polito.it}
\author{Gabriele Goletto\corref{cor}}\ead{gabriele.goletto@polito.it}
\author{Mirco Planamente}\ead{mirco.planamente@polito.it}
\author{Andrea Bottino}\ead{andrea.bottino@polito.it}
\author{Barbara Caputo}\ead{barbara.caputo@polito.it}
\author{Giuseppe Averta}\ead{giuseppe.averta@polito.it}
\fntext[fn1]{Corresponding author.}
\address{Department of Control and Computer Engineering, Politecnico di Torino\\Corso Castelfidardo, 34/d, Turin, 10138, Italy}
%\affiliation[1]{organization={Department of Control and Computer Engineering, Polytechnic of Turin},

%% Abstract
\begin{abstract}
Human activities exhibit a strong correlation between actions and the places where these are performed, such as washing something at a sink. 
More specifically, in daily living environments we may identify particular locations, hereinafter named \textit{activity-centric zones}, which may afford a set of homogeneous actions. 
Their knowledge can serve as a prior to favor vision models to recognize human activities.
However, the appearance of these zones is scene-specific, limiting the transferability of this prior information to unfamiliar areas and domains. This problem is particularly relevant in egocentric vision, where the environment takes up most of the image, making it even more difficult to separate the action from the context.
In this paper, we discuss the importance of decoupling the domain-specific appearance of activity-centric zones from their universal, domain-agnostic representations, and show how the latter can improve the cross-domain transferability of Egocentric Action Recognition (EAR) models. 
We validate our solution on the \epickitchens and Argo1M datasets.
Project page: \href{https://gabrielegoletto.github.io/EgoZAR/}{gabrielegoletto.github.io/EgoZAR}.

\end{abstract}

%% Keywords

\end{frontmatter}

\section{Introduction}
\label{sec:intro}
The privileged perspective offered by egocentric vision has proven highly effective in tracking human activities in daily life, thanks to the camera constantly following the wearer \citep{plizzari2023outlook, goletto2023bringing}. 
While providing an advantageous viewpoint on ongoing activities, the first-person perspective also brings the background remarkably close to the camera, inherently increasing its prominence in the field of view compared to third-person videos.

In this context, the concept of environmental affordance plays a pivotal role in connecting the wearer's activity with the underlying physical space. 
Specifically, the notion of \textit{affordance} has been extensively studied in neuroscience and cognitive psychology since the seminal work of~\cite{gibson2014ecological}. 
Affordances describe the potential actions or uses suggested by the physical characteristics of objects or the surrounding environment. This concept has recently gained attention in egocentric vision \citep{mur2023multi, luo2022learning}.
In particular, the work of~\cite{nagarajan2020ego} refers to environmental affordances as \textit{activity-centric zones}, defined as spatial locations, affording a coherent set of interactions, \eg a sink or a stove in a kitchen. 
\begin{figure}[t]
\centering
\vspace{\baselineskip}
\vspace{\baselineskip}
\includegraphics[width=.95\columnwidth]{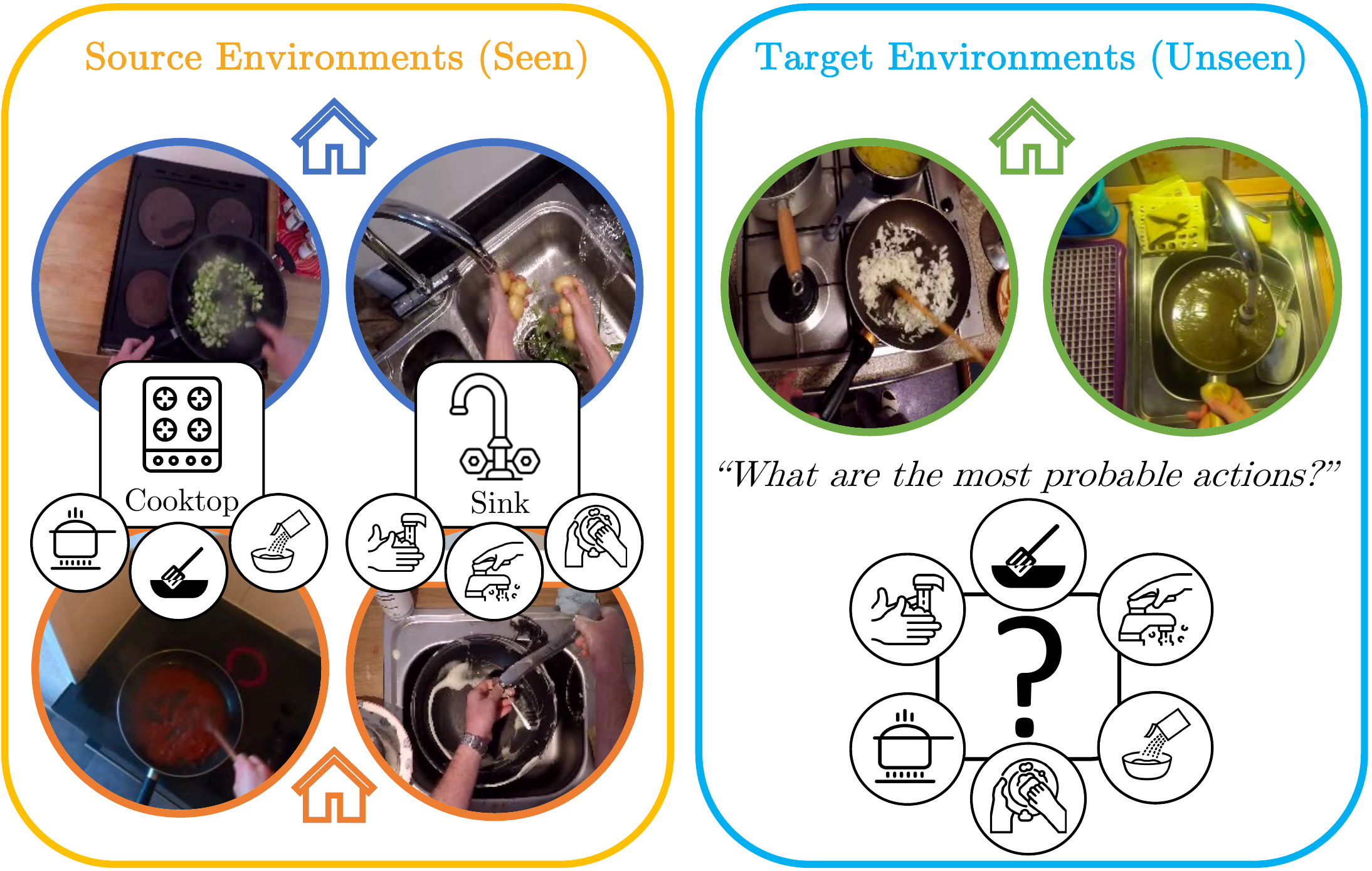}
\caption{
The actions a person performs in a scene are closely related to the specific places where they are performed (\textit{environmental affordances} \citep{nagarajan2020ego}). 
Current egocentric action recognition models learn these correlations during training, but struggle when faced with an unfamiliar environment, losing context.}
\label{fig:teaser}
\end{figure}

This prompts us to explore whether and how activity-centric zones are currently exploited for egocentric video understanding models, connecting human actions with the persistent underlying environment.
In particular, we demonstrate that the co-occurrence of specific actions in certain locations, predominantly present in egocentric vision, leads action recognition models to naturally learn a relationship between the actions and the locations in which they occur, exploiting it for making context-aware predictions.

This phenomenon, observable in the training data, is commonly known in computer vision as \textit{co-occurrence bias}~\citep{singh2020don} and, in our case, it aids the model in identifying a limited set of potential actions based on what is visible in the camera's field of view. For instance, when a user looks at a sink, it is more likely that the action being performed is washing rather than cooking. This process mirrors how people, in their daily lives, use their understanding of objects and tools to navigate unfamiliar environments and identify the activities the environment can afford.

However, despite their ability to exploit activity-centric locations observed during training, we demonstrate that current Egocentric Action Recognition (EAR) networks lack a mechanism to explicitly model the contribution of the environment in their inference process on unobserved zones. 
In other words, the co-occurrence bias, which aids the model in autonomously learning environmental affordances, leads to confusion in predictions as soon as the appearance of the zone changes.

Indeed, extensive egocentric vision datasets, such as \ek~\citep{damen2020rescaling} and Ego4D~\citep{grauman2022ego4d}, have a number of actions (i.e. verb-noun combinations) greatly exceeding the number of environments in which they were recorded.
As a consequence, models trained on these datasets overly depend on appearance-based features, such as visual representations of objects and tools, for recognizing actions~\citep{singh2020don}. This reliance leads the models to effectively recognize activity-centric zones only when tested with data from the same training environments, struggling to exploit the environmental affordances in new domains (Fig.~\ref{fig:teaser}).

To address this issue, this work focuses on universal representations of the activity-centric zones, which - we show - have the potential to assist RGB models in removing domain-specific biases from the encoding of activity-centric zones. 
By leveraging a domain-agnostic representations of these locations, we aim to isolate the domain-specific representation of the activity-centric zones in EAR models and replace them with a more general, domain-independent equivalent, resulting in more general EAR models. The main goal of our work is to address two key questions: \textit{how can we detect and identify these locations in real world conditions?}
And, \textit{can we use a domain-agnostic representation of these locations to improve the generalization capability of first person  action recognition models?}

We evaluate our approach on the EPIC-Kitchens-100 (\ek)~\citep{damen2020rescaling} and Argo1M~\citep{Plizzari_Perrett_Caputo_Damen} datasets in a Domain Generalization (DG) setting, where multiple source domains are available at training time but no target data can be accessed.
In summary, this paper presents the following contributions:
\begin{itemize}
    \item we shed light on the side-effect of the co-occurrence bias in egocentric video processing, which steer models in indirectly learning domain-dependent information about the environments (\ie domain-specific activity-centric zones);
    \item we propose \ours, an architecture which adopts more general representations of activity-centric zones to improve action recognition performance on unseen domains, enabling models to leverage the \textit{environmental affordances} even in unknown zones;
    \item we demonstrate with extensive experiments on the \ek and Argo1M datasets how replacing domain-specific environmental representations with their universal counterparts can help action recognition on unseen environments, achieving state-of-the-art Domain Generalization results on \ek and competitive performance on Argo1M.
\end{itemize}

\begin{figure*}[ht!]
     \centering
     \hfill
     \begin{subfigure}[b]{0.495\textwidth}
         \centering
         \includegraphics[width=0.975\textwidth,trim=0.5cm 1.75cm 0.5cm 2.0cm]{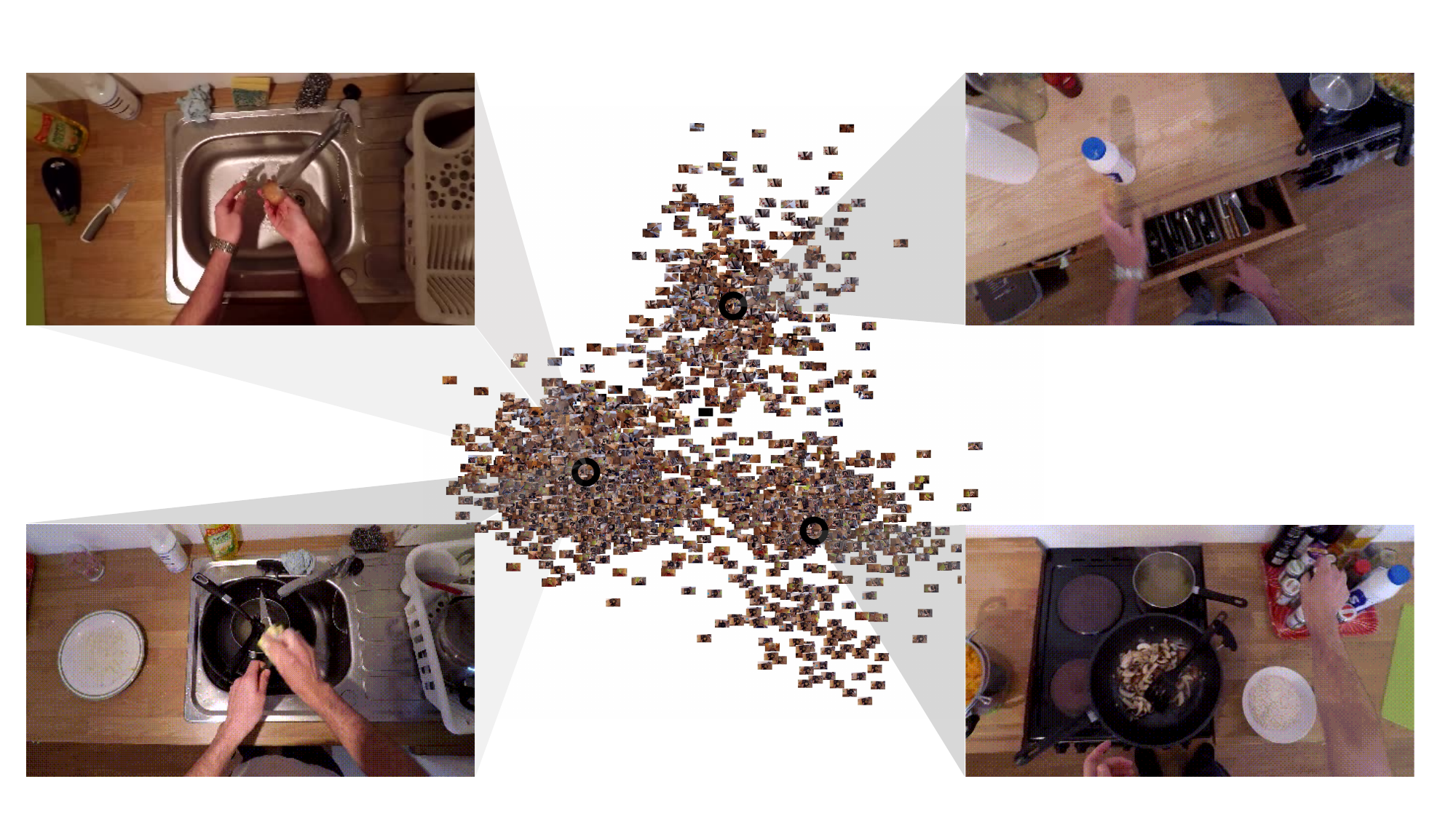}
         \caption{Source domain}
         \label{fig:clusters_src}
     \end{subfigure}
     \hfill
     \begin{subfigure}[b]{0.495\textwidth}
         \centering
         \includegraphics[width=0.975\textwidth,trim=0.5cm 1.75cm 0.5cm 2.0cm]{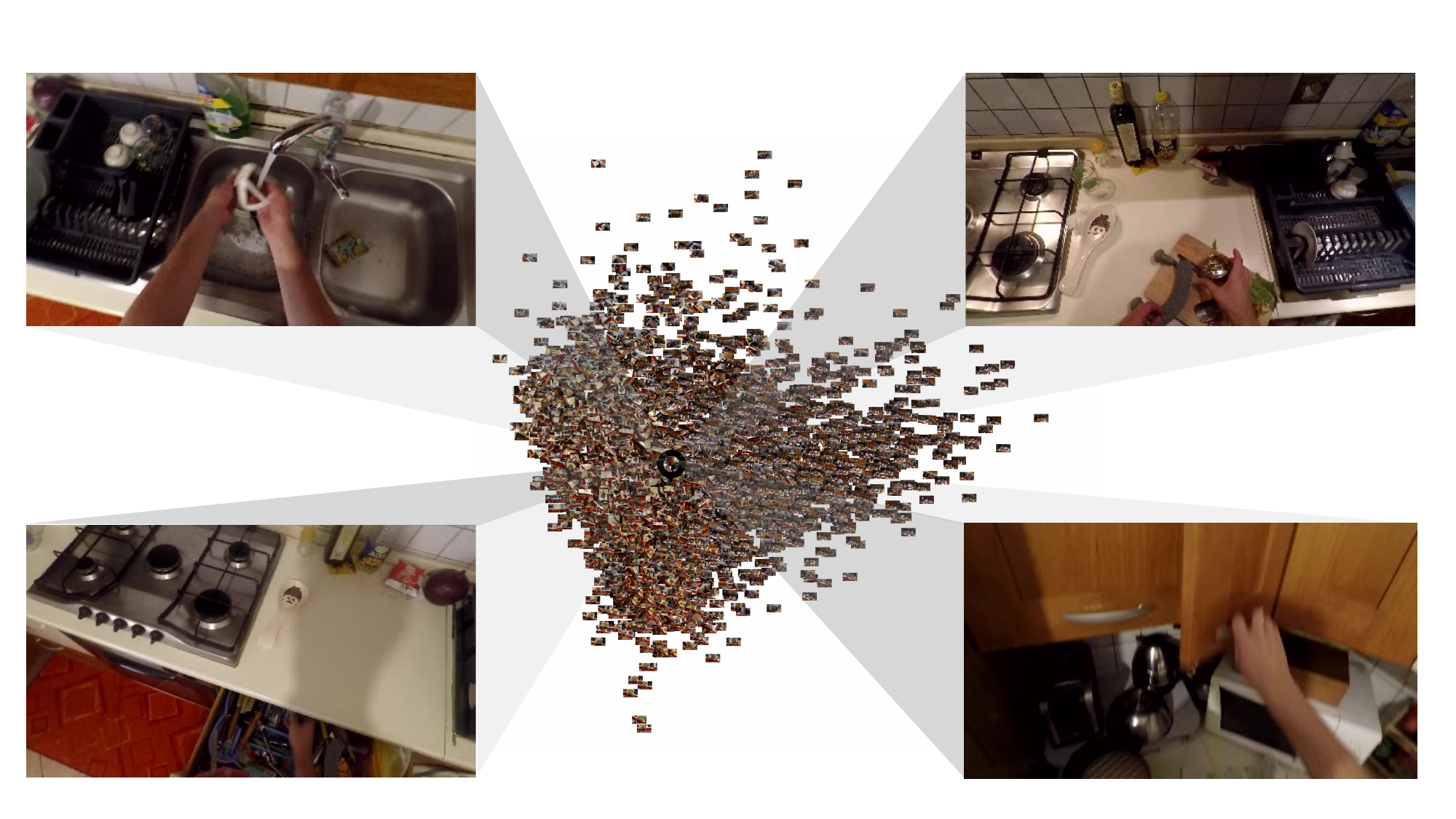}
         \caption{Target domain}
         \label{fig:clusters_tgt}
     \end{subfigure}
     \hfill
    \caption{Feature space of an EAR model. On the left (Fig.~\ref{fig:clusters_src}), the features obtained from a model trained and tested in the same environment are well separated based on the location where the actions are taking place. On the right (Fig.~\ref{fig:clusters_tgt}), when the same model is used in a different environment, this clustering effect is not present anymore and different locations are mapped to the same region of the feature space.}
    \vspace{.25cm}
\label{fig:clusters}
\end{figure*}

\section{Related works}
\label{sec:related}

\paragraph{Objects Affordances and activity-centric zones}
James J. Gibson defined the term \textit{affordances} in 1979 in the field of cognitive psychology~\citep{gibson2014ecological} referring to the physical properties of an object (or environment) that support certain human actions and interactions. 
The concept of affordance is now being widely explored in computer vision~\citep{hassanin2021visual,nagarajan2019grounded}, robotic manipulation~\citep{ardon2019learning,hamalainen2019affordance} and navigation~\citep{wang2020affordance}, and human-computer interaction~\citep{kaptelinin2012affordances}.
Most of the previous works on the topic focus on human-object interactions~\citep{yu2023fine, luo2022learning, kjellstrom2011visual}, object grasping~\citep{mandikal2021learning} and affordance detection \citep{do2018affordancenet}.
The concept of affordances has also been recently generalized to scenes.
Most notably, EGO-TOPO~\citep{nagarajan2020ego} extracts environmental affordances from egocentric videos and builds a topological map of the locations of the environment.
These \textit{so-called} activity-centric zones represent the main spatial regions in which actions may occur, driving interest towards their use in action recognition. 
More recently, \cite{mur2023multi} built EPIC-Aff, a dataset based on \ek providing multi-label pixel-wise affordance annotations with the camera pose. 

\paragraph{Egocentric Action Recognition}
Action recognition is one of the most studied tasks in egocentric vision~\citep{plizzari2023outlook}. 
The first architectures used in this context usually come from the third-person literature and fall into the categories of 2D CNN-based methods~\citep{10.5555/2968826.2968890, lin2019tsm} and 3D CNN-based methods~\citep{carreira2017quo, feichtenhofer2019slowfast}. 
LSTM and its variants \citep{Sudhakaran_2017_ICCV, planamente2021self} followed this first wave to better encode temporal information.
The most popular technique is the multi-modal approach \citep{Munro_2020_CVPR, furnari2020rolling}, especially in \ek competitions \citep{damen2020rescaling}, to combine the complementary information provided by different modalities, e.g. RGB and optical flow.
However, although optical flow has proven to be a strong modality for the action recognition task, it is computationally expensive. 
As shown in~\cite{Crasto_2019_CVPR}, the use of optical flow limits the application of several methods in online scenarios, pushing the community either towards single-stream architectures~\citep{zhao2019dance,planamente2021self}, or to investigate alternative modalities~\citep{kazakos2021slow,plizzari2022e2}. 

\paragraph{Video Domain Adaptation}
The goal of Unsupervised Domain Adaptation (UDA) is to close the gap between a labeled source domain and an unlabeled target domain. This task has been studied in detail in the context of image classification~\citep{long2015learning, ganin2016domain}.
UDA for video analysis has been primarily focused on extending existing techniques to include the temporal dimension~\citep{chen2019temporal} and/or the multi-modal nature of videos~\citep{Munro_2020_CVPR}.

Unlike UDA, the goal of DG~\citep{wang2022generalizing} is to improve generalization to out-of-distribution data without requiring access to the target data, using only data from one or more source training domains.
DG has been studied in different contexts from object recognition~\citep{koniusz2017domain}, to semantic segmentation~\citep{cordts2016cityscapes} and face recognition~\citep{shi2020towards}.
Applications to video are more scarce. 
Among these, RNA-Net~\citep{planamente2024relative} improves modalities cooperation on unseen scenarios by aligning feature norms. VideoDG~\citep{yao2021videodg} learns to align the local temporal features across different domains. CIR~\citep{Plizzari_Perrett_Caputo_Damen} reconstructs samples from different domains to learn more domain-agnostic representations.
Unlike previous approaches, ours is the first to emphasize the importance of activity-centric zones in improving domain generalization.

\section{Proposed method}
\label{sec:method}
Activity-centric zones provide useful insights into which actions are most likely to occur at a given location in the environment. However, exploiting these insights across different domains is not straightforward and requires models to reason about the location while ignoring its appearance.
We provide more intuitions behind this behavior in Sec.~\ref{sec:method_motivation}. 
We describe how to extract domain-agnostic representations for activity-centric zones in Sec.~\ref{sec:method_extraction} and show how these features can be integrated in an action recognition pipeline in Sec.~\ref{sec:method_integration}.

\begin{figure}[t]
\centering
\vspace{2.5mm}
\includegraphics[width=\linewidth]{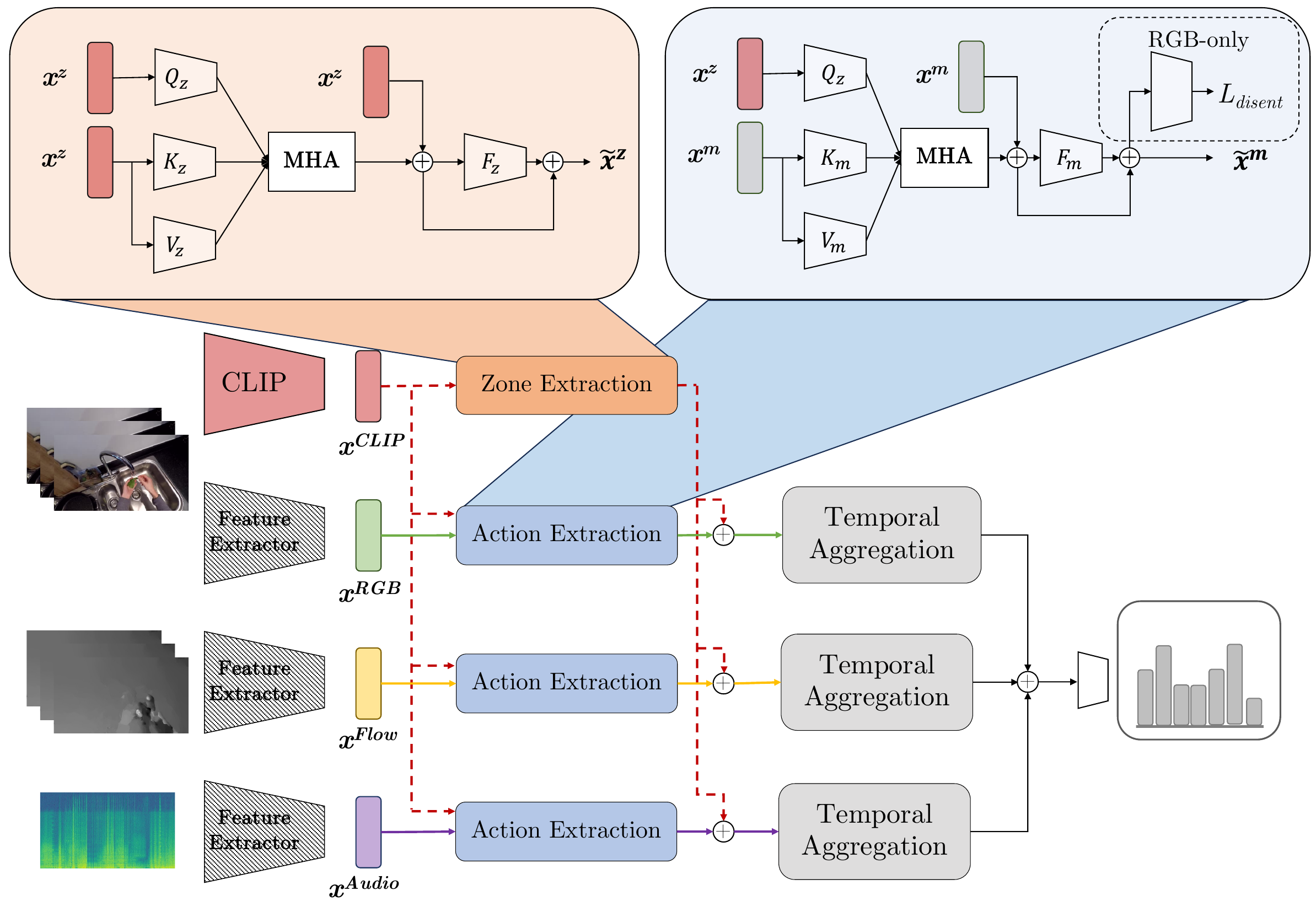}
\caption{
    Architecture of \ours.
    Each modality $m$ is processed using a separate features extractor to obtain the features $x^m$. CLIP features are then adopted both in the Zone and Action Extraction modules. This approach helps the network to focus on the activity-centric zones while minimizing the impact of environmental bias from the training domains. After temporal aggregation, the contributions from the different modalities $y_M$ are combined to produce the final prediction of the action label.
}
\label{fig:arch}
\end{figure}

\subsection{Intuition}\label{sec:method_motivation}
In egocentric vision, cameras are often positioned very close to the actions and the surrounding environments, causing RGB models to focus strongly on the environment.
We observe empirical evidence of this phenomenon looking at the feature space of an EAR model (Fig.~\ref{fig:clusters}).
Actions occurring in the same activity-centric zone are mapped to the same region of the feature space, regardless of their action label, which suggests that the model has learned a positive correlation between the environment and the set of actions that can be performed at a given location.
This clustering in feature space highlights the importance of the environment for action recognition.
However, this phenomenon does not transfer easily to new domains, as activity-centric zones are strongly coupled to their appearance, and models struggle to recognize the former (the semantic of the location) while ignoring the latter (its appearance).
Indeed, comparing the feature spaces on the train and test data, which belong to different visual domains, reveals that the clustering by activity-centric zones is no longer present when evaluating test data, as shown on the right side of Fig.~\ref{fig:clusters}.
To overcome this limitation, it is essential to allow the model to learn these activity-centric priors without the domain appearance bias, which is inherently present in video datasets and results from limited variability in the number of environments and locations represented.
This would allow models to embed this prior knowledge about the distribution of actions in a given location while avoiding the negative influence of domain-specific biases that hinder generalization.
Based on these observations, we identify two main challenges. 
First, the inclusion of domain-agnostic representations of the activity-centric zones into EAR models. 
Second, the development of a strategy for training an action-recognition model that uses these domain-agnostic representations to leverage the contextual information provided by the environment.

\subsection{Extracting activity-centric zone features}\label{sec:method_extraction}
We propose a method that leverages visual-language models trained on large-scale image datasets 
as a \textit{zone recognition model} to detect the activity-centric zones from the video stream. 
Indeed, being trained on millions of (image, caption) pairs sourced from the internet, these models are intrinsically able to recognize and generate similar features for the same location, \eg a sink or a stove in a kitchen, across different environments.
We use the features obtained from the zone recognition model to i) include a domain-agnostic information from the environment, and ii) remove the environment information from the input features of the action recognition model.
We adopt an unsupervised clustering algorithm on the features of the zone recognition model to discover clusters in the features space that correspond to different locations in which the actions occur.

Given a dataset of egocentric human actions, we define each sample $x_i$ as a triplet $x_i = (\mathbf{x}_i^m, \mathbf{x}_i^z, y_i)$, where $\mathbf{x}_i^m \in \mathbb{R}^{N\,\times\,D_m}$ and $\mathbf{x}_i^z \in \mathbb{R}^{N\,\times\,D_z}$ represent respectively the features extracted from an action recognition model $\mathcal{M}$ and a zone recognition model $\mathcal{Z}$ from $N$ uniformly sampled clips across the input video segment. 
Additional implementation details on the features extraction processes are presented in Sec.~\ref{sec:exp_setup}. 
Zone features from all the samples in the training dataset $\mathbf{x}_i^s$ are averaged over the clip dimension and clustered using K-Means in the euclidean features space.
This results in a set of $K$ prototypes that represent the centers of the clusters $\mathbf{c}_k \in \mathbb{R}^{D_z}$, each corresponding to a different location.
During training, each sample is assigned to the closest cluster using euclidean distance to obtain the corresponding activity-centric zone pseudo-label $y_i^z = \min_{k} ||\mathbf{x}_i^z - \mathbf{c}_k||_2$.

\subsection{Integration of the activity-centric zones}\label{sec:method_integration}
To integrate the prior information provided by the zone recognition model we propose \ours (see Fig.~\ref{fig:arch}). 
Our proposed architecture introduces two attention-based modules, namely the Zone Extraction (ZE) and Action Extraction (AE) modules, to explicitly separate the input features into two components, encoding zone and motion clues respectively.
The ZE module extracts the relevant zone-related information from the zone features $\mathbf{x}_i^z$, while the AE module encourages the action recognition features $\mathbf{x}_i^m$ to ignore the zone and domain appearance biases they incorporate. 
These modules are implemented using Multi-Head Attention followed by a linear projection and a residual connection.
Queries are computed from the zone features while keys and values are obtained from the zone or action features for the ZE and AE modules respectively.
Formally, the updated features $\mathbf{\tilde{x}}_i^z$ and $\mathbf{\tilde{x}}_i^m$ are computed as follows:
\begin{align}
\mathbf{o}_i^z = \mathbf{x}_i^z + \sigma\left(\frac{Q_z(\mathbf{x}_i^z)K_z(\mathbf{x}_i^z)^T}{\sqrt{D_z}}\right) \cdot V_z(\mathbf{x}_i^z), \hspace{.25cm} &\mathbf{\tilde{x}}_i^z = \mathbf{o}_i^z + F_z(\mathbf{o}_i^z), \\
\mathbf{o}_i^m = \mathbf{x}_i^m + \sigma\left(\frac{Q_m(\mathbf{x}_i^z)K_m(\mathbf{x}_i^m)^T}{\sqrt{D_z}}\right) \cdot V_m(\mathbf{x}_i^m), \hspace{.25cm} &\mathbf{\tilde{x}}_i^m = \mathbf{o}_i^m + F_m(\mathbf{o}_i^m),
\end{align}
where $Q$, $K$ and $V$ represent the queries, keys and values projections of the features and $F$ is a linear projection.
Then, the updated features are concatenated on the clips dimension and fed to a TRN~\citep{zhou2018temporal} layer:
\begin{equation}
    \mathbf{x}_i = TRN\left( [\mathbf{x}_i^z, \, \mathbf{x}_i^m]\right),
\end{equation}
where $\mathbf{x}_i \in \mathbb{R}^{D_m}$ and TRN is implemented as a linear projection, followed by a Batch Normalization layer, a ReLU activation and a dropout layer. Finally, features $\mathbf{x}_i$ are fed to a linear classifier that outputs the action logits $\mathbf{\tilde{y}}_i$.

\paragraph{Disentanglement of the action features}\label{sec:method_disent}
The objective of the AE module is to leverage the clues brought by the zone features to remove the appearance component of the motion features. 
To encourage this behavior, we introduce an adversarial classifier on top of the output of the AE module $\mathbf{\tilde{x}}_i^m$. 
The objective of the classifier is to recognize the activity-centric zone from the motion features. As a result, these features are pushed to discard any residual zone information. %, with the help of the AE module. 
The classifier is implemented as a two-layers MLP with hidden size 256, Batch Normalization and ReLU activations. The classifier outputs the activity-centric zone logits $\mathbf{\tilde{y}}^d_i$.

\subsection{Training and inference}\label{sec:method_training}
\ours architecture is trained jointly using Cross Entropy loss on the action logits $\mathbf{\tilde{y}}_i$ and on the output of the adversarial activity-centric zone classifier $\mathbf{\tilde{y}}^d_i$ with the supervision of the zone pseudo-labels. 
Other modalities, such as optical flow and audio, are less impacted by environmental bias, even though they can still benefit from the contextual features extracted from the zone recognition model.
As an example, an audio model aware of its proximity to a sink can more easily understand if sounds are linked to activities like washing, leveraging contextual clues for inference.
When training with multiple input modalities, the network is replicated for each modality and the \textit{modality-specific} action logits are averaged before computing the loss. 
In this context, the network is trained with a double Cross Entropy loss on both the fused (averaged) logits as well as on the \textit{modality-specific} logits. The Disentanglement Cross Entropy loss is computed just on the RGB modality, as it is the modality most affected by domain appearance biases. 

\section{Experiments and results}
\subsection{Experimental Setup}\label{sec:exp_setup}
\begin{figure*}[ht]
    \centering
    \includegraphics[width=0.98\textwidth]{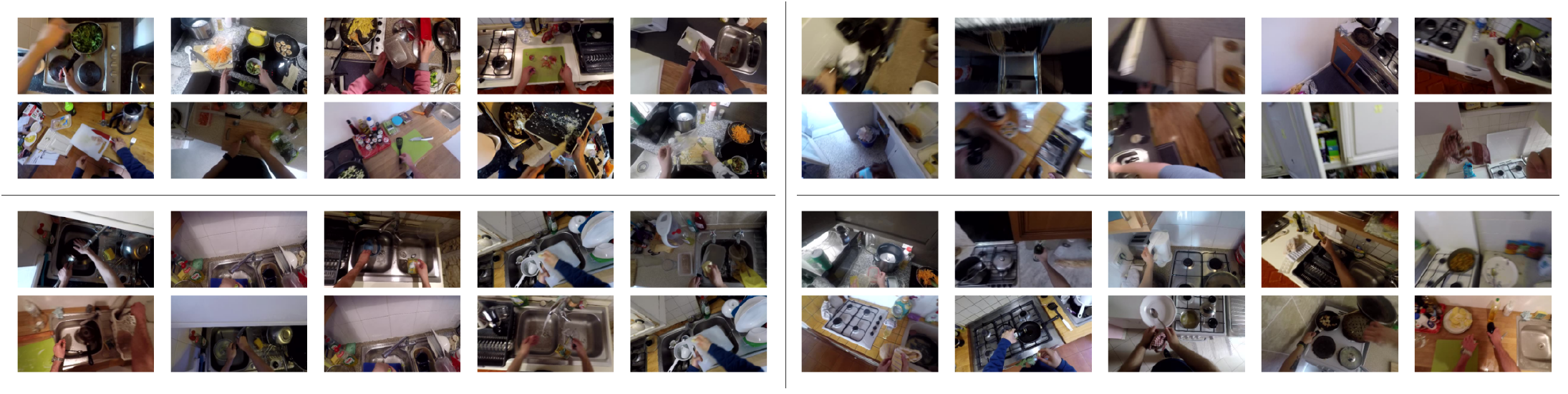}
    \caption{Clusters obtained with K-Means (K=4) on the CLIP features of \ek, showing how the same locations, for example sinks and stoves, but different kitchens are clustered together.}
    \label{fig:ek100_clusters}
\end{figure*}

\begin{figure}[ht]
    \centering
    \includegraphics[width=0.98\columnwidth]{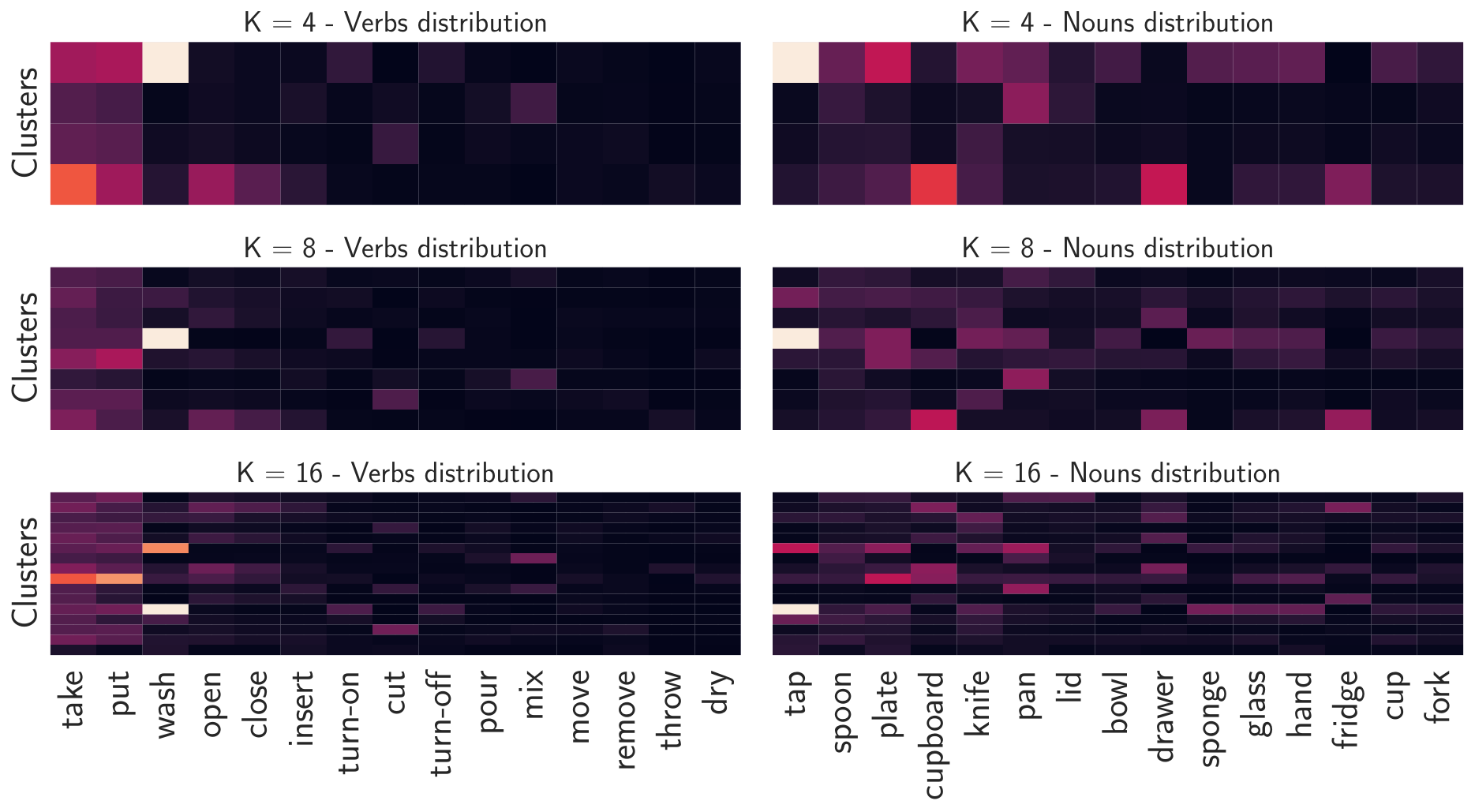}
    \caption{Cluster-wise verbs and nouns distributions showing quite distinct functional dependencies between clusters and the corresponding actions. Lighter colors indicate higher density.}
    \label{fig:vn-dist}
\end{figure}

\paragraph{Dataset}
We evaluate \ours on the Unsupervised Domain Adaptation (UDA) benchmark subset from \ek~\citep{damen2020rescaling}, a large dataset of fine-grained activities in a kitchen environment.
The dataset includes two forms of domain shift: i) each participant records its actions in a different kitchen (\textit{location shift}) and ii) the source and target splits partially share environments, although they are separated by a time interval (\textit{time shift}).
Our approach focuses on Domain Generalization (DG), thus the target split is not used during the training process.
Each action is annotated using a \textit{(verb, noun)} pair from a combination of 97 verb classes and 300 noun classes. 
Performances are reported using Top-1 and Top-5 accuracies for verbs, nouns and actions on the target validation set of \ek.
We also evaluate \ours on Argo1M~\citep{Plizzari_Perrett_Caputo_Damen}, a large scale egocentric vision dataset for Domain Generalization across different scenarios and locations. Argo1M consists of 10 splits, in each of which a specific scenario and location are not seen during training and only used for evaluation. Performance are reported using Top-1 Accuracy.

\paragraph{Implementation and Training Details}
For EK100, RGB, optical flow and audio features are extracted using the TBN architecture~\citep{Kazakos_2019_ICCV} finetuned on the source train split on EK100, following the protocol described in~\cite{damen2020rescaling}.
The projection layers and the attention modules are trained for 30 epochs, using the SGD optimizer with weight decay $1e-5$ and momentum 0.9. The learning rate is initially set to $1e-3$ and reduced by a factor 0.1 after epochs 10 and 20.
For the zone recognition model, we adopt various variants of CLIP~\citep{radford2021learning} and SWAG~\citep{singh2022revisiting}.

For Argo1M, we reuse the same hyperparameters as CIR~\citep{Plizzari_Perrett_Caputo_Damen} and learning rate $1e-6$. RGB and zone features are extracted using SlowFast~\citep{feichtenhofer2019slowfast} and CLIP ViT-L/14 respectively.
The features extractors and the zone recognition models are not updated during the training process.

\subsection{Analysis of unsupervised environment clustering}
Our approach identifies the locations in which actions are being performed through an unsupervised clustering of the features extracted with the zone recognition model.
Figure~\ref{fig:ek100_clusters} shows samples from clusters derived from the EK100 dataset. The data was clustered using K-Means with a value of K set to 4, employing the L2 distance metric on features extracted with CLIP ViT-L/14.
We observe that functionally similar locations are naturally clustered together in the CLIP's features space.
Additionally, we integrate these qualitative observations with the \textit{per-cluster} verbs and nouns distributions, computed for different number of clusters, as shown in Figure~\ref{fig:vn-dist}.
The plot confirms the presence of functional dependencies between the clusters and the labels distributions, with the exception of some verbs, \eg, \textit{take} and \textit{put} that are not tied to specific locations.

\subsection{\ek Results}\label{sec:ek_results} 
We present a comparison of state-of-the-art methods on EK100 in Table~\ref{tab:ek100_main}, comparing \ours with MM-SADA~\citep{Munro_2020_CVPR}, Gradient Blending~\citep{wang2020makes}, RNA~\citep{planamente2024relative} and CIR~\citep{Plizzari_Perrett_Caputo_Damen} in the DG setting and with TA3N~\citep{chen2019temporal} and CIA~\citep{yang2022interact} in UDA.
To account for variations in the network architectures used by these models and ensure a fair comparison, we report each model with its \textit{Source Only} performance, corresponding to standard training using cross-entropy only.
This dataset poses significant challenges, with limited improvements and difficulties in comparing the results of different methods, e.g. \textit{Source Only} of Gradient Blending outperforms TA3N or MM-SADA.
Comparison of UDA and DG approaches leads to similar observations, as the gap between the best approaches in these two settings is quite small.
We also evaluate CIR on \ek, using the same \textit{Source Only} as \ours. 
CIR benefits from the detailed narrations in Argo1M, the dataset for which it was originally proposed, while \ek narrations are less descriptive, mostly a simple concatenation of the verb and noun labels, limiting its performance.

Our solution shows significant improvements over the previous SOTA without access to target data. 
We observe noticeable gains especially in action and noun accuracy, indicating that \ours enables better reasoning about the manipulated objects and their interactions in key locations across environments.
Indeed, the domain-agnostic clues introduced by the zone features reduce the negative effect of appearance biases, focusing less on the environment and helping the model in recognizing the same objects in different domains. 
Overall, \ours achieves a considerable improvement over the previous SOTA, without access to the target data.
\begin{table*}[ht]
\scriptsize
\centering
\caption{Results on the validation set of EPIC-Kitchens-100 dataset in Unsupervised Domain Adaptation (UDA) and Domain Generalization (DG) settings using RGB, Optical Flow and Audio.
To ensure a fair comparison we report the Source Only performance for each method. 
Our results are averaged over three runs. Best in \textbf{bold}. Second best is \underline{underlined}. 
$^\dagger$Reproduced.}
\vspace{-1mm}

\begin{tabularx}{\textwidth}{Xccc|ccc|ccc|c}
\toprule

& & & & \multicolumn{3}{c|}{\textbf{Top-1 Accuracy (\%)}} & \multicolumn{3}{c|}{\textbf{Top-5 Accuracy (\%)}} & \\

\textbf{Method} & \textbf{Modality} & \textbf{EAR Network} & \textbf{Setting} & \textbf{Action} & \textbf{Verb} & \textbf{Noun} & \textbf{Action} & \textbf{Verb} & \textbf{Noun} & \textbf{Mean Accuracy (\%)} \\
\midrule

Source Only & RGB-Flow-Audio & TBN-TRN & -  & 19.20 & 46.70 & 27.78 & 42.12 & 75.42 & 48.27 & 38.95\\ 
TA3N~\cite{chen2019temporal} & RGB-Flow-Audio & TBN-TRN & UDA  & 19.61 & 48.44 & 28.87 & 43.36 & 75.95 & 50.12 & 41.25 (\up $+2.30\%$) \\ 
\midrule

Source Only & RGB-Flow-Audio & TBN-TRN & - &18.99	&47.14	&27.35	&41.82	&75.27	&49.36 & 43.32\\
MM-SADA~\cite{Munro_2020_CVPR} & RGB-Flow-Audio & TBN-TRN & DG &19.15	&47.76	&27.93	&42.90	&77.07	&49.77 & 44.10  (\up $+0.78\%$)\\
MM-SADA~\cite{Munro_2020_CVPR} & RGB-Flow-Audio & TBN-TRN & UDA &19.25	&48.44	&28.26	&43.41	&77.56	&50.59 & 44.59  (\up $+1.27\%$)\\
\midrule

Source Only & RGB-Flow-Audio & TBN-TRN & - & 18.29 & 46.79 & 26.79 & 41.36 & 75.39 & 48.44 & 42.84\\
RNA~\cite{planamente2024relative} & RGB-Flow-Audio & TBN-TRN & DG  & 19.81 & \underline{50.75} & 27.92 & 46.76 & 80.64 & 51.37 & 46.21 (\up $+3.37\%$)\\
RNA~\cite{planamente2024relative} &RGB-Flow-Audio & TBN-TRN & UDA  & 20.05 & \textbf{50.82} & 29.19 &   46.04 & 80.89 & 52.18 & 46.53 (\up $+3.69\%$)\\
\midrule

Source Only & RGB-Flow-Audio & TBN-TRN & - & 19.61 & 47.69 & 28.48 & - & - & - & - \\ 
CIA~\cite{yang2022interact}& RGB-Flow-Audio & TBN-TRN & UDA & 20.30 & 48.34 & 29.50 & - & - & - & - \\ 

\midrule

Source Only & RGB-Flow-Audio & TBN-TRN & DG & 19.96 & 50.27 & 29.04 & 46.74 & \underline{81.74} & 52.14 & 46.65\\
Gradient Blending~\cite{wang2020makes} & RGB-Flow-Audio & TBN-TRN & DG & 20.26 & 50.18 & \underline{29.60} & 46.86 & \textbf{81.82} & 52.57 & 46.88 (\up $+0.23\%)$ \\

\midrule
\midrule

Source Only & RGB-Flow-Audio & TBN-TRN & - & 19.41 & 49.09 & 29.17 & 45.89 & 80.72 & 52.42 & 46.16 \\

CIR~\cite{Plizzari_Perrett_Caputo_Damen} (w/o text)$^\dagger$ & RGB-Flow-Audio & TBN-TRN & DG & 19.41 & 49.45 & 29.13 & 46.82 & 80.64 & 53.49 & 46.49 (\up $+0.33\%$) \\

CIR~\cite{Plizzari_Perrett_Caputo_Damen}$^\dagger$ & RGB-Flow-Audio & TBN-TRN & DG & 19.43 & 48.82 & 29.08 & 46.94 & 81.07 & 53.25 & 46.43 (\up $+0.27\%$) \\
\midrule

Source Only & RGB-Flow-Audio & TBN-TRN & - & 19.41 & 49.09 & 29.17 & 45.89 & 80.72 & 52.42 & 46.16 \\

\textbf{\ours (RN50)} & RGB-Flow-Audio & TBN-TRN & DG & \underline{20.32} & 50.05 & 29.53 & \underline{46.95} & 81.18 & \underline{53.65} & \underline{46.95} (\up $+0.79\%$) \\
\textbf{\ours (ViT-L/14)} & RGB-Flow-Audio & TBN-TRN & DG & \textbf{21.83} & 50.41 & \textbf{31.99} & \textbf{50.06} & 81.27 & \textbf{58.13} & \textbf{48.95} (\up $+2.79\%$) \\

\bottomrule
\end{tabularx}
\label{tab:ek100_main}
\end{table*}

\begin{table}[ht]
\caption{Contribution of the attention and disentanglement components of \ours across different input modalities.}
\vspace{-1mm}
\scriptsize
\centering
\setlength{\tabcolsep}{5pt}

\begin{tabularx}{\columnwidth}{X|ccc|ccc|c}
\toprule

& \multicolumn{3}{c|}{\textbf{Top-1 Accuracy (\%)}} & \multicolumn{3}{c|}{\textbf{Top-5 Accuracy (\%)}} & \textbf{Mean}\\

& \textbf{Action} & \textbf{Verb} & \textbf{Noun} & \textbf{Action} & \textbf{Verb} & \textbf{Noun} & \textbf{Acc. (\%)} \\
\midrule

RGB & 10.91 & 33.76 & 21.80 & 36.97 & 75.40 & 43.72 & 37.09 \\
+ Attn. & 13.17 & 36.13 & 24.32 & 41.76 & 76.71 & 49.22 & 40.22 \\
+ Disent. & \textbf{13.63} & \textbf{37.33} & \textbf{25.06} & \textbf{42.46} & \textbf{77.18} & \textbf{50.34} & \textbf{41.00} \\

\midrule

Flow & 13.05 & 44.69 & 20.57 & 35.51 & 77.44 & 40.50 & 38.63 \\
+ Attn. & \textbf{16.80} & \textbf{46.02} & \textbf{25.92} & \textbf{43.97} & \textbf{79.00} & \textbf{51.37} & \textbf{43.85} \\
\midrule

Audio & 8.18 & 32.36 & 13.78 & 27.07 & 70.47 & 31.89 & 30.63 \\
+ Attn. & \textbf{14.97} & \textbf{39.74} & \textbf{23.65} & \textbf{40.34} & \textbf{75.90} & \textbf{47.99} & \textbf{40.43} \\

\bottomrule
\end{tabularx}
\label{tab:ek100_sm}
\end{table}

\paragraph{Single modality training}
The integration of the zone information may be also beneficial for modalities that lack visual clues of the environment. These modalities suffer less from environmental biases but can benefit from the integration of zones information.
We show the effect of the integration of the AE and SE modules in unimodal AR models in Table~\ref{tab:ek100_sm}, observing a significant improvement compared to the baselines, especially on the noun metric.
For RGB, the modality most affected by visual domain bias, we report results using both the attention and disentanglement modules of \ours.
Compared to the baseline, we observe an overall improvement of $+3.13\%$ using the attention modules and $+3.91\%$ when also the disentanglement loss is introduced. 
Notably, the better Top-1 verb accuracy indicates that the network is leveraging the contextual and domain-agnostic location clues provided by CLIP features to identify the action being performed.

\paragraph{Comparison of different zone recognition models}
\begin{table}[t]
\caption{Comparison of models for zone features extraction using the cross attention component. Experiments conducted with RGB only.}
\scriptsize
\centering
\setlength{\tabcolsep}{5pt}

\begin{tabularx}{\columnwidth}{X|ccc|ccc|c}
\toprule

& \multicolumn{3}{c|}{\textbf{Top-1 Acc. (\%)}} & \multicolumn{3}{c|}{\textbf{Top-5 Acc. (\%)}} & \textbf{Mean}\\

\textbf{Arch.} & \textbf{Action} & \textbf{Verb} & \textbf{Noun} & \textbf{Action} & \textbf{Verb} & \textbf{Noun} & \textbf{Acc. (\%)} \\

\midrule

Baseline & 10.91 & 33.76 & 21.79 & 36.97 & 75.40 & 43.72 & 37.09 \\

\midrule
\multicolumn{8}{c}{SWAG~\cite{singh2022revisiting}}\\
\midrule

ViT/B-16 & 11.60 & 34.24 & 22.58 & 39.35 & 75.44 & 46.80 & 38.34 \\
ViT/L-16 & 12.06 & 34.24 & 23.64 & 40.19 & 75.59 & 48.10 & 38.97 \\

\midrule
\multicolumn{8}{c}{CLIP~\cite{radford2021learning}}\\
\midrule

RN-50 & 11.69 & 35.23 & 22.43 & 38.54 & 75.69 & 45.56 & 38.19 \\
ViT-B/32  & 11.56 & 34.38 & 22.44 & 38.26 & 75.00 & 45.51 & 37.86 \\
ViT-B/16  & 11.88 & 34.82 & 22.87 & 39.30 & 75.23 & 46.73 & 38.47 \\
ViT-L/14  & \textbf{13.17} & \textbf{36.13} & \textbf{24.32} & \textbf{41.76} & \textbf{76.71} & \textbf{49.22} & \textbf{40.22} \\

\bottomrule
\end{tabularx}
\label{tab:scene_models}
\vspace{2.5mm}
\end{table}

\begin{table}[ht!]
\caption{Ablation on the number of clusters for disentanglement of unimodal RGB model.}
\vspace{-1mm}
\scriptsize
\centering

\setlength{\tabcolsep}{5pt}

\begin{tabularx}{\columnwidth}{X|ccc|ccc|c}
\toprule

& \multicolumn{3}{c|}{\textbf{Top-1 Accuracy (\%)}} & \multicolumn{3}{c|}{\textbf{Top-5 Accuracy (\%)}} & \textbf{Mean}\\

\textbf{K} & \textbf{Action} & \textbf{Verb} & \textbf{Noun} & \textbf{Action} & \textbf{Verb} & \textbf{Noun} & \textbf{Acc. (\%)} \\

\midrule

- & 13.17 & 36.13 & 24.32 & 41.76 & 76.71 & 49.22 & 40.22 \\

\midrule

2 & \textbf{13.71} & 37.10 & 24.97 & 41.96 & 77.04 & 49.62 & 40.73 \\ 
4 & 13.63 & \textbf{37.33} & \textbf{25.06} & \textbf{42.46} & 77.18 & \textbf{50.34} & \textbf{41.00} \\ 
8 & 13.44 & 37.12 & 24.37 & 42.30 & 76.76 & 49.92 & 40.65 \\ 
16 & 13.08 & 36.68 & 24.12 & 42.00 & 76.86 & 49.57 & 40.38 \\ 
32 & 13.29 & \textbf{37.33} & 24.59 & 42.33 & \textbf{77.36} & 49.89 & 40.80 \\ 

\bottomrule
\end{tabularx}
\label{tab:ek100_k}
\vspace{-2.5mm}
\end{table}
We evaluate in Table~\ref{tab:scene_models} the impact of different backbones for features extraction, using multiple CLIP and SWAG~\citep{singh2022revisiting} variants. The latter is a ViT architecture trained for image classification using weak supervision of hashtags and is the current SOTA for scene classification on Places-365~\citep{zhou2014learning}, suggesting it could be useful in our context to recognize the activity-centric zones.
Even the least capable model (CLIP RN50) considerably outperforms the baseline, proving the effectiveness of the attention modules of \ours, and larger models consistently provide higher average accuracy. 
Additionally, \ours shows robust performance improvements using different zone recognition models.
We attribute the better performance of CLIP compared to SWAG to the former being trained on more descriptive captions using an image-language contrastive objective, compared to the weak supervision of the hashtags used in the training process of SWAG.

\newcommand{\India}{\includegraphics[scale=0.16]{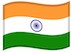}}
\newcommand{\Building}{\includegraphics[scale=0.035]{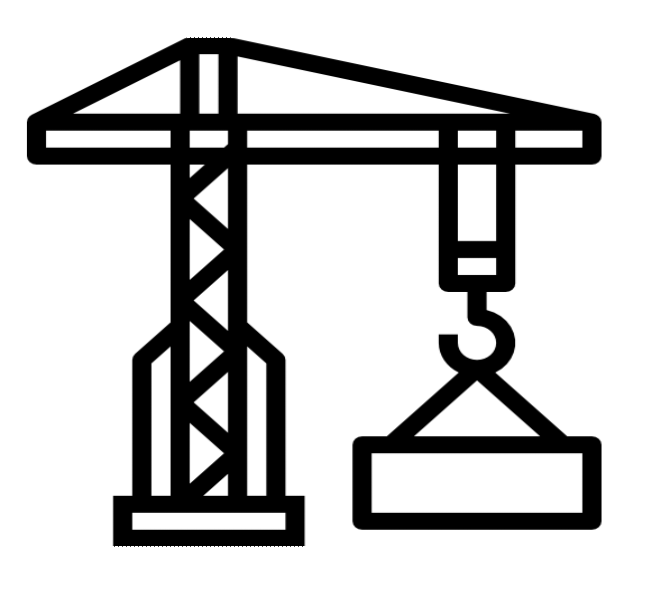}}
\newcommand{\Saudi}{\includegraphics[scale=0.14]{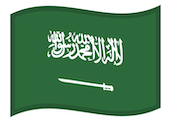}}
\newcommand{\US}{\includegraphics[scale=0.14]{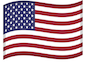}}
\newcommand{\Knitting}{\includegraphics[scale=0.65]{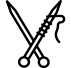}}
\newcommand{\Sport}{\includegraphics[scale=0.65]{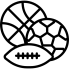}}
\newcommand{\Shopping}{\includegraphics[scale=0.75]{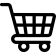}}
\newcommand{\Cleaning}{\includegraphics[scale=0.85]{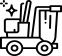}}
\newcommand{\Playing}{\includegraphics[scale=0.65]{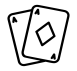}}
\newcommand{\Gardening}{\includegraphics[scale=0.75]{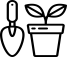}}
\newcommand{\Japan}{\includegraphics[scale=0.14]{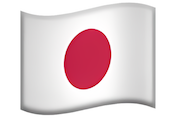}}
\newcommand{\Italy}{\includegraphics[scale=0.14]{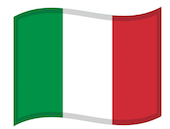}}
\newcommand{\Colombia}{\includegraphics[scale=0.14]{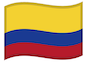}}
\newcommand{\Sewing}{\includegraphics[scale=0.025]{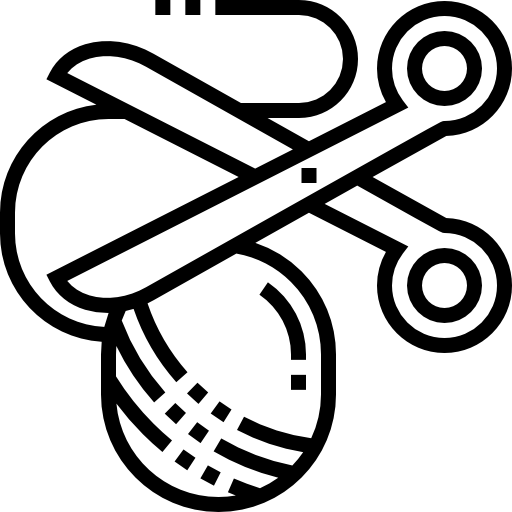}}%
\newcommand{\Mechanic}{\includegraphics[scale=0.025]{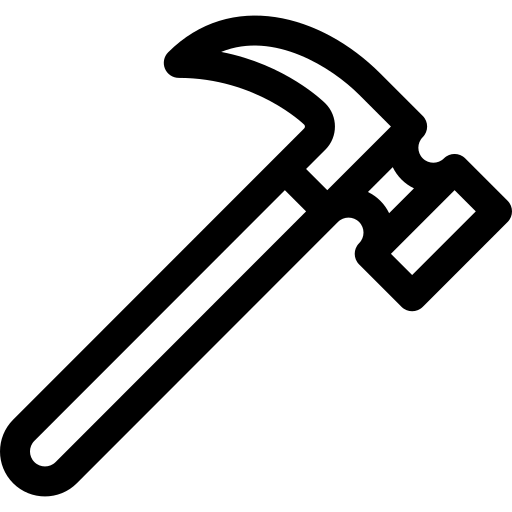}}%
\newcommand{\Cooking}{\includegraphics[scale=0.025]{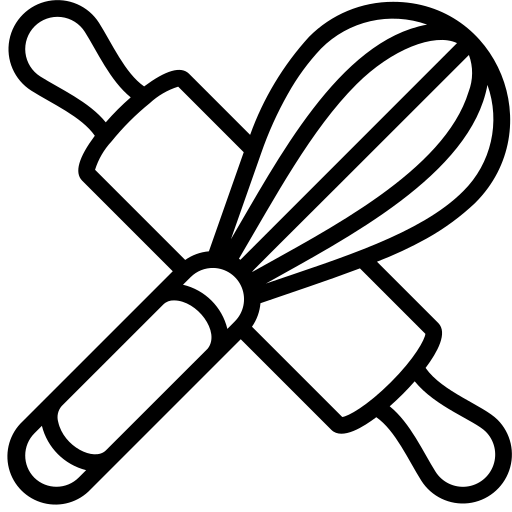}}%

\setlength{\tabcolsep}{3.5pt}
\begin{table*}[t]
\caption{Top-1 accuracy on ARGO1M~\citep{Plizzari_Perrett_Caputo_Damen}. Best results in \textbf{bold}, second best {underlined}. $^\dagger$: Domain labels required during training. \textit{D}: distribution matching, \textit{A}: adversarial learning, \textit{M}: label-wise mix-up, \textit{P}: domain-prompts, \textit{R}: reconstruction, \textit{T}: video-text association, \textit{Z}: activity-centric zone learning.}
\vspace{-1mm}
\label{tab:argo1m}
\centering
\scriptsize
\begin{tabularx}{\textwidth}{Xcccccccccccccccccc} 
\toprule
\multicolumn{1}{l}{} &\multicolumn{7}{c}{DG Strategies} & \Gardening \US & \Cleaning \US & \Knitting \India & \Shopping \India & \Building \US & \Mechanic \Saudi & \Sport \Colombia & \Cooking \Japan & \Sewing \Italy & \Playing \US & \\ \cline{2-8}
\multicolumn{1}{l}{} & & & & & & & & & & & \multicolumn{1}{c}{} & \multicolumn{1}{c}{} &\multicolumn{1}{c}{} & \multicolumn{1}{c}{} & \multicolumn{1}{c}{} &\\
\multicolumn{1}{l}{\multirow{-3}{*}{}} &\multirow{-2}{*}{D} &\multirow{-2}{*}{A} &\multirow{-2}{*}{M} &\multirow{-2}{*}{P} &\multirow{-2}{*}{R} &\multirow{-2}{*}{T}&\multirow{-2}{*}{Z} &\multirow{-2}{*}{\textbf{\begin{tabular}[c]{@{}c@{}}Ga\\ US-PNA\end{tabular}}} &\multirow{-2}{*}{\textbf{\begin{tabular}[c]{@{}c@{}}Cl\\ US-MN\end{tabular}}} &\multirow{-2}{*}{\textbf{\begin{tabular}[c]{@{}c@{}}Kn\\ IND\end{tabular}}}&\multirow{-2}{*}{\textbf{\begin{tabular}[c]{@{}c@{}}Sh\\ IND\end{tabular}}} &\multirow{-2}{*}{\textbf{\begin{tabular}[c]{@{}c@{}}Bu\\ US-PNA\end{tabular}}}&\multirow{-2}{*}{\textbf{\begin{tabular}[c]{@{}c@{}}Me\\ SAU\end{tabular}}} &\multicolumn{1}{c}{\multirow{-2}{*}{\textbf{\begin{tabular}[c]{@{}c@{}}Sp\\ COL\end{tabular}}}} &\multicolumn{1}{c}{\multirow{-2}{*}{\textbf{\begin{tabular}[c]{@{}c@{}}Co\\ JPN\end{tabular}}}} &\multicolumn{1}{c}{\multirow{-2}{*}{\textbf{\begin{tabular}[c]{@{}c@{}}Ar\\ ITA\end{tabular}}}} &\multicolumn{1}{c}{\multirow{-2}{*}{\textbf{\begin{tabular}[c]{@{}c@{}}Pl\\ US-IN\end{tabular}}}} &\multirow{-3}{*}{\textbf{Mean}} \\ 
\midrule
Random & & & & & & & & \textcolor{white}{0}{\color[HTML]{8d8d8d}} 8.00 & 10.64 & \textcolor{white}{0}{\color[HTML]{8d8d8d}}9.13 & 14.36 & \textcolor{white}{0}{\color[HTML]{8d8d8d}}9.55 & 13.04 & \textcolor{white}{0}{\color[HTML]{8d8d8d}}8.35 & 10.13 & \textcolor{white}{0}{\color[HTML]{8d8d8d}}9.86 & 15.68 &{\cellcolor[HTML]{EFEFEF}}10.84 \\
ERM &&&&&&&&20.75 &22.35 &18.69 &22.14 &20.73 &23.51 &18.97 &24.81 &22.75 &23.29 &{\cellcolor[HTML]{EFEFEF}}21.80 \\
\midrule
CORAL$^\dagger$~\cite{sun2016deep} &\cmark &&&&&&&22.14 &22.55 &19.07 &24.01 &22.18 &24.31 &19.16 &25.36 &23.89 &{25.96} &{\cellcolor[HTML]{EFEFEF}22.86} \\
DANN$^\dagger$~\cite{ganin2016domain} &\cmark &\cmark &&&&&& {22.42} &{23.85} &19.27 &22.89 &22.23 &23.70 &18.64 &25.86 &23.86 &23.28 &{\cellcolor[HTML]{EFEFEF}}22.60 \\
MMD$^\dagger$~\cite{li2018domain} &\cmark &&&&&&&{22.42} &23.60 &19.66 & 24.46 &22.08 &24.64 & {19.59} &25.87 &23.84 &24.78 &{\cellcolor[HTML]{EFEFEF}}{23.09} \\
Mixup~\cite{wang2020heterogeneous} &&&\cmark &&&&&21.97 &22.21 &{19.90} &23.81 &21.45 &24.35 &19.01 & 25.90 &23.85 &24.41 &{\cellcolor[HTML]{EFEFEF}}22.69 \\
BoDA$^\dagger$\cite{yang2022multi} &\cmark &&&&&&&22.17 &22.78 &19.62 &22.94 &21.46 &23.97 &19.18 &25.68 &23.92 &24.90 &{\cellcolor[HTML]{EFEFEF}}22.66 \\
DoPrompt$^\dagger$~\cite{zheng2022prompt} &&&&\cmark &&&&21.92 &22.77 & {20.40} &23.67 & {22.75} & {24.67} &18.24 &25.04 & {24.74} & 25.24 &{\cellcolor[HTML]{EFEFEF}}22.94 \\
CIR w/o text~\cite{Plizzari_Perrett_Caputo_Damen} &&&&& \cmark &&&23.39	&24.52&\underline{21.02} &\underline{26.62}	&\underline{24.64}	&\textbf{27.00}	&\underline{19.66}	&25.42	&\underline{25.71}	&\underline{30.17} &{\cellcolor[HTML]{EFEFEF}}\underline{24.81} \\ 
CIR~\cite{Plizzari_Perrett_Caputo_Damen} &&&&&\cmark &\cmark && \underline{24.10} & \underline{25.51} &20.46 &\textbf{27.78} &\textbf{24.93} &\underline{26.83} &\textbf{19.75} &\textbf{26.34} &25.67 &\textbf{30.94} &{\cellcolor[HTML]{EFEFEF}}\textbf{25.23} \\ 

\midrule

\textbf{\ours$^\dagger$} &&&&&&&\cmark& \textbf{24.53} & \textbf{26.12} & \textbf{21.70} & 25.82 & {24.05} & {24.88} & 18.91 & \underline{26.02} & \textbf{26.05} & {29.94} & {\cellcolor[HTML]{EFEFEF}}\underline{24.80} \\

\bottomrule 
\end{tabularx}
\end{table*}

\paragraph{Ablation on the number of clusters}

We analyze in Table~\ref{tab:ek100_k}, the impact of different number of clusters.
All configurations exceed the performance of attention modules alone and we observe similar performances across a large set of values. 
We attribute this behavior to two factors. 
First, the number of locations in EK100 is limited and mostly dominated by sinks and stoves, which occur frequently. 
Second, having more clusters means that the larger clusters are broken into smaller chunks, although the disentanglement objective, which encourages the network to become more confused about the locations, remains the same. 
Unless otherwise specified, we set $K=4$ for all disentanglement experiments. 

\subsection{ARGO1M Results}
Argo1M~\citep{Plizzari_Perrett_Caputo_Damen} features actions from a collection of different scenarios (\eg, \textit{Cooking} and \textit{Sport}) and locations (\eg, \textit{United States} and \textit{India}).
While certain scenarios, such as \textit{cooking} or \textit{cleaning}, benefit from EgoZAR's activity-centric zones, others, such as \textit{shopping}, are less suitable due to the even distribution of the same actions across different locations in the environment.
The heterogeneity in Argo1M's data distribution required some minor adjustments to the clustering process adopted in \ours.
Indeed, activity-centric zones are typically associated with the scenario, \eg a sink and an oven in the kitchen, but the location can introduce confounding factors.
For example, kitchens in the USA can differ significantly from those in Saudi Arabia. To account for this, we clustered zone features separately by location and then merged clusters from different locations based on similarity of action distributions. This approach clusters zones that are visually distinct but support similar actions and may represent the same activity-centric zone.

Despite these limitations, Argo1M can be considered the largest and most diverse setting for DG in egocentric vision and a valuable addition to our analysis (Table~\ref{tab:argo1m}).
\ours outperforms all previous methods and is \textit{on-par} with CIR~\citep{Plizzari_Perrett_Caputo_Damen}, which was specifically designed to deal with the many different scenarios and locations of Argo1M.
CIR recombines samples from different scenarios and locations to learn a more agnostic representation that does not depend on the context in which the action occurs. On the contrary, we argue that the role of the environment is crucial in action recognition across different domains, and therefore build \ours to leverage the location information while reducing the impact of the appearance bias.
The advantage of \ours compared to CIR is more evident in settings like \ek where environmental affordances are more dominant, as discussed in Sec.~\ref{sec:ek_results}.

\section{Limitations and future works}
\ours heavily relies on the assumption that human activities are highly correlated with the locations in which they occur.
This behavior is very evident in datasets like \ek, and more noisy in other datasets such as Argo1M. This may impact the applicability of \ours to new datasets and partially reduce its effectiveness. Also, our approach requires to tune the number of clusters which is a very \textit{dataset-dependent} parameter.
Future works could focus on making the clustering process more flexible to discover the activity-centric zones in a completely unsupervised way, without using any prior knowledge on the number of clusters.

\section{Conclusions}
In this paper, we showcase the impact of environmental affordances in action recognition. We argue that the leveraging this environmental information is significantly influenced by their appearance, strongly limiting the generalization ability to other areas and domains. 
We propose \ours, a method to exploit zone-recognition models as source of a domain-agnostic information on the activity-centric zones where the actions are taking place, and to replace domain-specific appearance of activity-centric zones.
Extensive experiments on \ek show the effectiveness of \ours, achieving SOTA performance and highlighting how the integration of zone information may help in action recognition.

\section*{Acknowledgements}
This study was supported in part by the CINI Consortium through the VIDESEC project and carried out within the FAIR - Future Artificial Intelligence Research and received funding from the European Union Next-GenerationEU (PIANO NAZIONALE DI RIPRESA E RESILIENZA (PNRR) – MISSIONE 4 COMPONENTE 2, INVESTIMENTO 1.3 – D.D. 1555 11/10/2022, PE00000013). This manuscript reflects only the authors’ views and opinions, neither the European Union nor the European Commission can be considered responsible for them. G. Goletto is supported by PON “Ricerca e Innovazione” 2014-2020 – DM 1061/2021 funds.

\bibliographystyle{model5-names}
\bibliography{refs}

\end{document}